\documentclass{article}

\usepackage{microtype}
\usepackage{graphicx}
\usepackage{booktabs} 

\usepackage{hyperref}

\usepackage[accepted]{icml2025}

\usepackage{amsmath}
\usepackage{amssymb}
\usepackage{mathtools}
\usepackage{amsthm}

\usepackage[capitalize,noabbrev]{cleveref}

\usepackage{setspace}
\usepackage{multirow}
\usepackage{tablefootnote}
\usepackage{amsfonts}
\usepackage{bbm}
\usepackage{caption}
\usepackage{subcaption}
\usepackage{bm}
\usepackage{float}
\usepackage{makecell}
\usepackage{wrapfig}
\usepackage{mathrsfs} 
\usepackage{adjustbox}

\theoremstyle{plain}

\theoremstyle{definition}

\theoremstyle{remark}

\usepackage[textsize=tiny]{todonotes}

\icmltitlerunning{X-IL: Exploring the Design Space of Imitation Learning Policies}

\begin{document}

\twocolumn[
\icmltitle{X-IL: Exploring the Design Space of Imitation Learning Policies}



\icmlsetsymbol{equal}{*}
\icmlsetsymbol{dagger}{$\dagger$}

\begin{icmlauthorlist}
\icmlauthor{Xiaogang Jia}{kit}
\icmlauthor{Atalay Donat}{kit}
\icmlauthor{Xi Huang}{kit}
\icmlauthor{Xuan Zhao}{ind}
\icmlauthor{Denis Blessing}{kit}
\icmlauthor{Hongyi Zhou}{kit}
\icmlauthor{Han A. Wang}{dagger,meta}
\icmlauthor{Hanyi Zhang}{uk}
\icmlauthor{Qian Wang}{kit}
\icmlauthor{Rudolf Lioutikov}{kit}
\icmlauthor{Gerhard Neumann}{kit}
\end{icmlauthorlist}

\icmlaffiliation{kit}{Karlsruhe Institute of Technology, Germany}
\icmlaffiliation{uk}{University of Liverpool}
\icmlaffiliation{ind}{Independent Researcher}
\icmlaffiliation{meta}{Reality Labs, Meta}

\icmlcorrespondingauthor{Xiaogang Jia}{jia266163@gmail.com}


\icmlkeywords{Imitation Learning, Robot Learning}

\vskip 0.3in
]



\printAffiliationsAndNotice{\icmlHanStatement} 

\begin{abstract}

Designing modern imitation learning (IL) policies requires making numerous decisions, including the selection of feature encoding, architecture, policy representation, and more. As the field rapidly advances, the range of available options continues to grow, creating a vast and largely unexplored design space for IL policies.
In this work, we present \textbf{X-IL}, an accessible open-source framework designed to systematically explore this design space. The framework's modular design enables seamless swapping of policy components, such as backbones (e.g., Transformer, Mamba, xLSTM) and policy optimization techniques (e.g., Score-matching, Flow-matching). This flexibility facilitates comprehensive experimentation and has led to the discovery of novel policy configurations that outperform existing methods on recent robot learning benchmarks. Our experiments demonstrate not only significant performance gains but also provide valuable insights into the strengths and weaknesses of various design choices. This study serves as both a practical reference for practitioners and a foundation for guiding future research in imitation learning. Code available at this  \href{https://github.com/ALRhub/X_IL}{link}.

\end{abstract}

\section{Introduction}
\begin{figure*}[t!]
    \centering
    \includegraphics[width=0.8\linewidth]{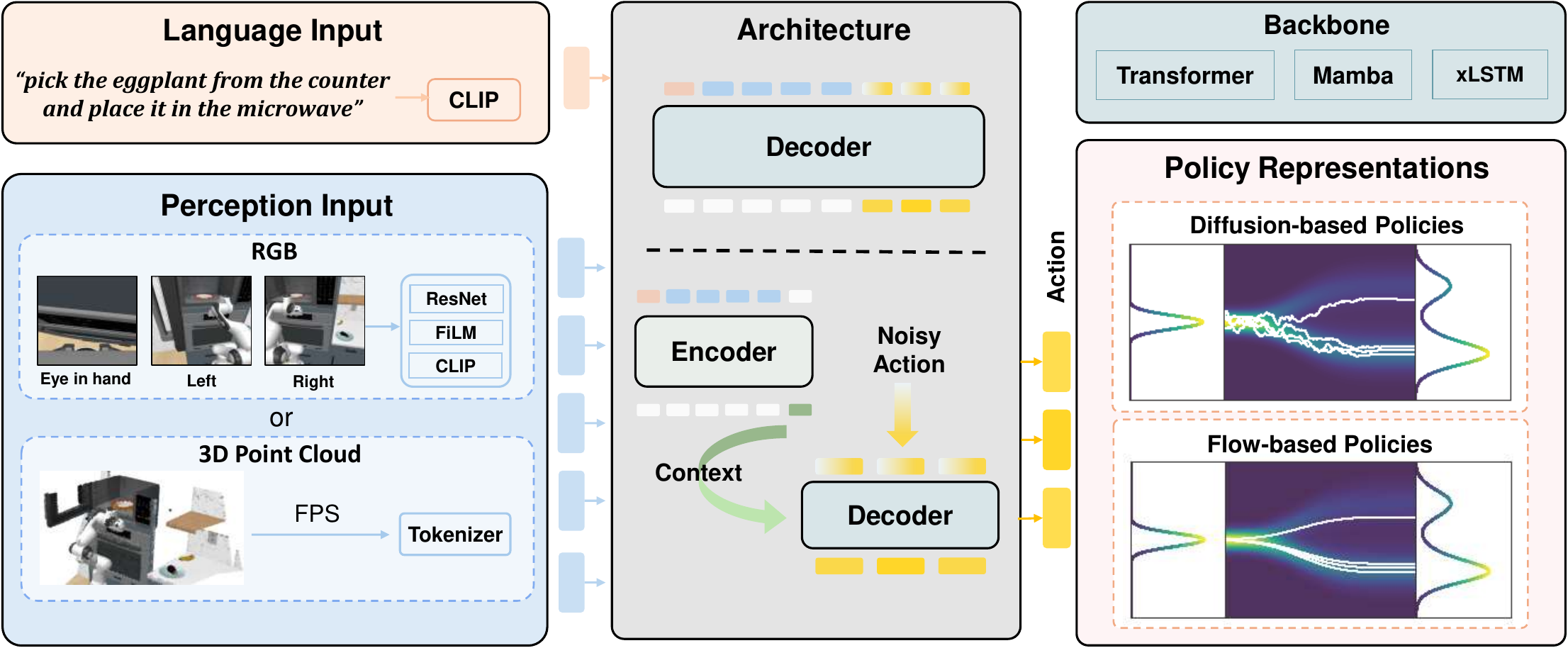}
    \caption{Overview of X-IL framework. X-IL supports multi-modal inputs (Language, RGB, and Point Cloud) and two architectures: Decoder-Only and Encoder-Decoder. Inside each architecture, the Backbone serves as the core computational unit, offering support for Transformer, Mamba, and xLSTM. For policy representations, X-IL supports Behavior Cloning (BC), Diffusion-based, and Flow-based Policies, enabling diverse learning paradigms for imitation learning. Notably, each component—input modality, architecture, backbone, and policy—can be easily swapped to efficiently explore various model configurations.
    }
    \label{fig:arch_moil}
\end{figure*}

Imitation learning (IL) \cite{osa2018algorithmic} has emerged as a powerful paradigm for teaching agents complex behaviors by mimicking demonstrations, eliminating the need for explicit reward engineering \cite{argall2009survey}. However, the rapid development of novel machine-learning techniques across various domains makes it challenging to assess their potential impact on imitation learning.
To address this, we introduce X-IL, a novel framework designed to integrate and explore recently developed techniques within an imitation learning pipeline. Our framework decomposes the imitation learning process into four key modules: (1) observation representations, (2) backbones, (3) architectures, and (4) policy representations. Each module is interchangeable, enabling systematic exploration of the design space for imitation learning policies. This modularity facilitates rapid prototyping, benchmarking, and deployment.
Figure \ref{fig:arch_moil} provides an overview of our framework.

More specifically, we offer various observation representations, including 2D RGB inputs and 3D point cloud representations, to accommodate diverse perception tasks. Our framework incorporates versatile encoders, such as MLPs, ResNet \cite{he2015deepresiduallearningimage}, ViT \cite{dosovitskiy2021imageworth16x16words}, and pre-trained models, which can be tailored to different input types. Furthermore, we offer both Decoder-only and Encoder-Decoder architectures. Decoder-only is simpler with fewer parameters, while Encoder-Decoder supports additional representation learning for better generalization and scaling.
We define the backbone as the core computational unit of the policy architecture, offering support for Transformer \cite{vaswani2017attention}, Mamba \cite{gu2024mambalineartimesequencemodeling}, and xLSTM \cite{beck2024xlstmextendedlongshortterm}.
Lastly, we offer several state-of-the-art policy representations including popular diffusion- and flow-based models \cite{ho2020denoising, chi2023diffusion, reuss2023goal,lipman2023flowmatchinggenerativemodeling, du2022toflowefficientcontinuousnormalizing}.


Our contributions can be summarized as follows:

\textbf{1)} We introduce X-IL, a user-friendly and highly modular framework for imitation learning that supports multi-modal inputs, flexible encoders, diverse architectures and policy representations.

\textbf{2)} We leverage our framework to systematically explore the design space of IL policies. In doing so, we obtain novel policy designs that achieve state-of-the-art results on the LIBERO \cite{liu2023liberobenchmarkingknowledgetransfer} and RoboCasa \cite{nasiriany2024robocasalargescalesimulationeveryday} benchmarks. 
 
\textbf{3)} Our extensive experiments yield valuable insights, such as the potential of recent sequence models as strong alternatives to Transformers, or, that fusing different input modalities can lead to significant performance improvements.


    

Our work is organized as follows. In \Cref{sec: related work}, we review related work. \Cref{sec: framework} describes the proposed framework, detailing design choices and its modular components. In \Cref{sec:experiments}, we provide experimental evaluations across multiple benchmarks, followed by a discussion of results and observations in Section \ref{sec: discussion}.

\section{Related Work} \label{sec: related work}

\textbf{Multi-modal Imitation Learning.} 
Early imitation learning methods relied on either state \cite{schaal1996learning, ho2016generative, torabi2018behavioral} or images \cite{pomerleau1988alvinn, lynch2020learning, young2021visual} to describe the environment and define the goal. However, obtaining accurate state information in real-world setting is not straightforward, and state-based representation struggles to capture the complexity of unstructured environments. Conversely, images provide a rich representation for behavior learning \cite{robomimic2021} and can be directly acquired from raw sensory input. Despite these advantages, using images as goal conditions in imitation learning is limited by their ambiguity in goal representation and difficulty in goal specification, making them less flexible for real-world deployment.
To address this, natural language has been explored as an alternative goal representation, offering a more intuitive and accessible way to specify tasks. Recent studies \cite{shridhar2022cliport,reuss2024multimodal, bharadhwaj2024roboagent} have explored the integration of language goals with image observation, enabling more flexible policy learning. Another line of research fine-tunes Vision-Language Models (VLMs) models to obtain Vision-Language Action Models (VLAs) \cite{kim2024openvlaopensourcevisionlanguageactionmodel, li2023vision, li2023generalist}. However, purely image-based representation lack crucial 3D structural information, which is essential for many tasks. Therefore, there is a recent trend on incorporating richer 3D scene presentations, such as point clouds, to enhance policy performance \cite{ke20243d, Ze2024DP3}. 

\textbf{Imitation Learning with Sequence Models.}
In recent years, sequence models have been increasingly applied to learning human behaviors, as human decision-making is inherently non-Markov and requires incorporating historical observations \cite{robomimic2021}. Early works utilized RNN-based structures\cite{robomimic2021}. However, these models suffers from vanish gradient for handling long observation sequence and low training efficient due to the sequential processing nature. To address this limitation, Transformer-based architectures have been widely adopted \cite{shafiullah2022behavior,reuss2023goal,bharadhwaj2024roboagent}, offering superior scalability and sequence modeling capabilities. Most recently, State-Space Models(SSM) \cite{gu2024mambalineartimesequencemodeling,jia2024mailimprovingimitationlearning} have emerged as a promising alternative to Transformers, demonstrating remarkable efficiency on small datasets and the ability to learn consistent representation.  Additionally, improved RNN-based architectures, such as xLSTM \cite{beck2024xlstmextendedlongshortterm}, have shown potential to rival both Transformer and SSMs in natural language processing. However, their application in imitation learning remains largely unexplored. X-IL aims to bridge this gap by incorporating Transformers, Mamba, and xLSTM as modular components and empirically evaluating their performance under different tasks and representations.

\textbf{Modular Imitation Learning Libraries.}
While numerous open-source libraries provide algorithm-specific implementation of imitation learning methods \cite{chi2023diffusion, lee2024behavior, jia2024mailimprovingimitationlearning}, only a few offer modular design that spans multiple algorithms and architectures. Robomimic \cite{robomimic2021} implements Behavior Cloning (BC) with MLP, RNN, and Transformer-based policies,
while Imitation \cite{gleave2022imitation} provides modular implementations of several imitation learning and inverse reinforcement learning methods. However, these libraries do not include recent diffusion-based imitation learning approaches.
To address this gap, a recent work, CleanDiffuser \cite{cleandiffuser}, introduces a modular implementation for diffusion models in decision-making, supporting policy architectures such as MLP, UNet, ResNet, and Transformer. However, its evaluation is limited to tasks with low dimensional state input and 2D image input. In contrast, X-IL expands modularity by supporting multi-modal inputs, including 2D images, point clouds, and language-conditioned goals. Additionally, X-IL integrates state-of-the-art sequence models, such as Mamba and xLSTM, broadening its applicability to more complex environments and diverse IL architectures.

\section{The X-IL Framework} \label{sec: framework}
In this section, we introduce X-IL, a modular and open-source framework for imitation learning, based on the following design principles:

{\textbf{Modularity.}} X-IL systematically decomposes the imitation learning pipeline into different modules with different components that are easily interchangeable. This modular design enables flexible integration and evaluation of different approaches, facilitating systematic exploration of the design space of imitation learning policies.

{\textbf{Ease-of-use principle.}} Our framework is easy to use, supporting popular tools such as Hydra \cite{Yadan2019Hydra} for configuration management and Weights \& Biases (Wandb) \cite{wandb} for logging and visualization, streamlining the experimentation process.

{\textbf{Incorporation of Novel Techniques.}} X-IL integrates recent advancements such as Mamba \cite{gu2024mambalineartimesequencemodeling} and xLSTM \cite{beck2024xlstmextendedlongshortterm} for sequence modeling and Diffusion and Flow Matching for policy learning, improving the efficiency and generalization of imitation learning policies.

To enable flexible experimentation, we decompose the imitation learning pipeline into four key modules: 1) observation representations, 2) backbones, 3) architectures, and 4) policy representations. An overview of our framework is shown in Figure \ref{fig:arch_moil}. Below, we provide a detailed description of each module and its components.

\subsection{Observation Representations} \label{sec: observation}
Our framework considers three primary types of representations: RGB inputs, Point Cloud, and Language.
Below, we introduce these representations and how we encode them.


\textbf{RGB Inputs.} 
Visual imitation learning has received significant attention in recent research \cite{chi2023diffusion}. RGB images, captured from multiple camera viewpoints, provide essential texture and semantic information for object recognition and scene understanding. Prior works have demonstrated that ResNet is a strong encoder for manipulation tasks, making it a widely adopted choice \cite{shafiullah2022behaviortransformerscloningk, wan2024lotuscontinualimitationlearning, zhu2023violaimitationlearningvisionbased}. To effectively leverage RGB data, X-IL supports various feature extractors, including ResNet, FiLM-ResNet \cite{turkoglu2022filmensembleprobabilisticdeeplearning}, and CLIP \cite{radford2021learningtransferablevisualmodels}, with a modular codebase that allows easy integration of additional image encoders. Each image is encoded as a single token and passed into the backbone (see \Cref{sec: backbones}) for further processing, enabling flexible and efficient representation learning.

\textbf{Point Cloud.}
Point clouds provide 3D spatial structures obtained from RGB-D cameras or LiDAR sensors, offering geometric information for manipulation tasks. Unlike RGB images, point clouds inherently encode object positions and shapes, making them ideal for tasks requiring fine-grained spatial reasoning. Prior works have emphasized the importance of preserving geometric features for effective representation learning \cite{wan2024lotuscontinualimitationlearning, ze20243ddiffusionpolicygeneralizable, gyenes2024pointpatchrlmaskedreconstruction}. In X-IL,
we use Furthest Point Sampling (FPS) \cite{qi2017pointnet} to downsample points, which helps preserve the geometric structures of the 3d space. We adopt a token-based representation to capture scene geometry efficiently. Our framework supports two encoders: a lightweight MLP with Max Pooling \cite{ze20243ddiffusionpolicygeneralizable} for computational efficiency and an attention-based encoder with class token for enhanced feature extraction. The implementation details can be found in Appendix \ref{subsec:repr-encoders}.

\textbf{Language.}
Language-guided imitation learning \cite{stepputtis2020languageconditionedimitationlearningrobot, lynch2021languageconditionedimitationlearning, mees2022matterslanguageconditionedrobotic, yu2023usingdemonstrationslanguageinstructions, reuss2024multimodaldiffusiontransformerlearning} has gained increasing attention as it provides a high-level, abstract way to describe tasks, object attributes, and robot actions, making it a valuable modality in imitation learning. Unlike visual and geometric inputs, language offers context that enhances generalization and adaptability across diverse tasks. To process language, X-IL integrates the pre-trained language model CLIP \cite{radford2021learningtransferablevisualmodels} to convert textual information into dense embeddings. These embeddings are then fused with visual and point cloud features, enabling a richer multimodal representation for policy learning.

\begin{figure}[h]
    \centering
    \includegraphics[width=0.4\linewidth]{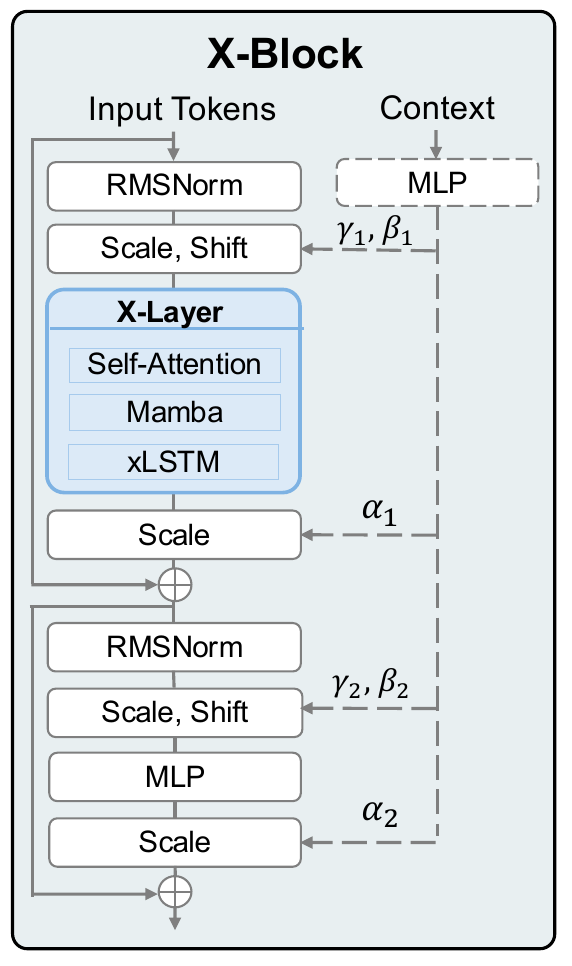}
    \caption{Network details of X-Block. X-Layer is the core part, which is used to process sequence tokens; AdaLn conditioning is used to inject the context information. Details can be found in Appendix \ref{subsec:x-block}.}
    \label{fig:x_block}
\end{figure}

\subsection{Backbones: X-Block} \label{sec: backbones}
The choice of backbone architectures is critical for learning effective policies in imitation learning, as it determines how different input modalities are processed and how sequential dependencies are captured. We define backbones as the core components responsible for modeling sequential information. Previous works \cite{chi2023diffusion, jang2022bczzeroshottaskgeneralization, chen2021decisiontransformerreinforcementlearning, rouxel2024flowmatchingimitationlearning, shaier2022datadrivenapproachespredictingspread, bertasius2021spacetimeattentionneedvideo} have predominantly utilized convolutional 1D architectures, such as U-Net \cite{ronneberger2015unetconvolutionalnetworksbiomedical}, or Transformer-based models for sequence modeling in imitation learning. However, recent advancements in sequence modeling, such as Mamba and xLSTM, remain largely unexplored in this domain. We provide a comparative analysis of these architectures:

\textbf{Transformer} \cite{vaswani2017attention}. A widely used attention-based model that has demonstrated strong performance in imitation learning due to its ability to handle non-Markovian behavior in human demonstrations. Most imitation learning models including Vision-Language Action Models (VLAs) use transformers as the backbone.

\textbf{Mamba} \cite{gu2024mambalineartimesequencemodeling}. A structured state-space model (SSM) that significantly improves the efficiency of SSMs while rivaling Transformers in performance. Unlike Transformers, Mamba maintains linear computational complexity. Mamba Imitation Learning (MaIL) \cite{jia2024mailimprovingimitationlearning} has shown that Mamba-based policies outperform Transformer-based policies with small datasets.

\textbf{xLSTM} \cite{beck2024xlstmextendedlongshortterm}. A variant of LSTM that is designed to enhance long-term dependency modeling while maintaining computational efficiency. Unlike standard LSTMs, which struggle with long-range dependencies, xLSTM incorporates architectural improvements to mitigate vanishing gradient issues. While recurrent models generally lack the expressiveness of self-attention, xLSTM offers a balance between efficiency and performance, making it a potential alternative for imitation learning tasks where computational constraints are a concern.

In X-IL, we aim to investigate the potential of these architectures for policy learning. Inspired by the DiT-Block \cite{peebles2023scalable} structure, our framework introduces X-Block, as illustrated in Figure \ref{fig:x_block}. The core component of X-Block is the X-Layer, which is responsible for processing temporal information. We provide three backbone options: Transformer, Mamba, and xLSTM, allowing flexibility in sequential modeling. Additionally, AdaLN conditioning \cite{peebles2023scalable} is incorporated—not only for conditioning time embeddings in diffusion models but also for integrating representation features. Our findings indicate that using representations as conditioning signals enhances performance, further improving the effectiveness of policy learning.

\subsection{Architectures} \label{sec: architectures}
The architecture of an imitation learning model defines how input representations are processed and how action outputs are generated. X-IL supports two architectures: Decoder-Only and Encoder-Decoder.
Prior works such as ACT \cite{zhao2023learning} and MDT \cite{reuss2024multimodaldiffusiontransformerlearning} adopt an encoder-decoder design, whereas PearceTransformer \cite{pearce2023imitating} and MoDE \cite{reuss2024efficient} follow a decoder-only approach.
Below, we introduce these architectures and explain their integration within our framework. The illustrations of them are given in Figure \ref{fig:arch_moil}. 

\textbf{Decoder-only Models.}
In X-IL, the Decoder-Only architecture is implemented by stacking multiple X-Blocks, where both observations and actions are jointly processed within the decoder. The model outputs only the action tokens, which are then used to train the policy representations.



\textbf{Encoder-Decoder Models.}
The Encoder-Decoder architecture in X-IL follows a two-stage approach: the Encoder first encodes multi-modal inputs into a latent representation, and the decoder then generates actions based on this structured embedding. 
Prior works primarily utilize cross-attention to connect the encoder’s output with the decoder’s input. However, Mamba and xLSTM lack a built-in mechanism to handle variable-length sequences in this manner. Instead, we find that AdaLN conditioning provides an efficient and flexible alternative for constructing the Encoder-Decoder architecture, enabling effective integration of encoded representations into the decoding process.






\subsection{Policy Representations} \label{sec: policy representations}
Besides naive behavior cloning approaches, our framework offers a variety of state-of-the-art policy representations, which can be broadly categorized as diffusion-based and flow-based models.

\textbf{Behavior Cloning} Behavior cloning (BC) assumes a Gaussian distribution as policy representation and maximizes the likelihood of predicted actions in the given ground truth distributions.


\textbf{Diffusion-Based Policies}
Denoising diffusion probabilistic models (DDPM) \cite{ho2020denoising} captures the score function field and iteratively optimizes the action. BESO \cite{reuss2023goal} is based on a continuous-time diffusion framework. BESO allows for varying diffusion steps, as well as diverse sampling techniques. Our framework supports both DDPM-style and continuous-time BESO-style policies.    

\textbf{Flow-Based Policies}  
Continuous-time normalizing flows trained via flow matching \cite{lipman2022flow} have recently gained a lot of attention and are also suitable as policy representations. 
These methods, often referred to as rectified flows (RF) \cite{liu2022flow} are fully supported in our framework.

More detailed description of the policy representations see Appendix \ref{appendix:policy}.

\section{Experiments}
\label{sec:experiments}

To explore the design space of Imitation Learning, we conduct extensive experiments on two robot learning benchmarks: LIBERO and RoboCasa. Our study systematically examines various backbones, architectures, and policy designs for both visual and point cloud-based imitation learning.


\subsection{Simulation Benchmark}

\textbf{LIBERO} \cite{liu2023liberobenchmarkingknowledgetransfer}
We evaluate our modular framework with various model architectures and policy heads using RGB inputs on the LIBERO benchmark, which comprises four distinct task suites: \textit{LIBERO-Spatial}, \textit{LIBERO-Object}, \textit{LIBERO-Goal}, and \textit{LIBERO-Long}. These task suites are specifically designed to evaluate different aspects of robotic learning and manipulation capabilities.

To thoroughly compare the performance of each architecture, we conduct evaluations using both 10 trajectories (20\% of the available demonstrations) and 50 trajectories (the full dataset). All models were trained for 100 epochs in LIBERO task suites, and we used the last checkpoint for evaluation. Following the official LIBERO benchmark settings, we simulated 50 rollouts for each sub-task, totaling 500 simulations per task suite. We report the average success rate for each task suite over 3 seeds. 

\begin{figure}[h]
    \centering
    \includegraphics[width=0.48\textwidth]{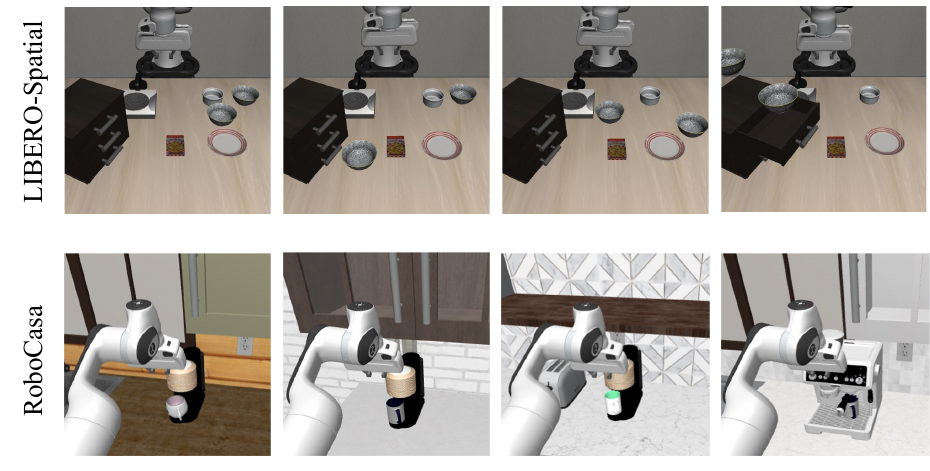}
    \caption{Illustration of LIBERO and RoboCasa. While LIBERO demonstrates minimal variations in the same task, e.g. LIBERO-Spatial, RoboCasa provides diversities in different aspects. CoffeeServeMug is shown in the figure.}
    \label{fig:libero_robocasa_diff}
\end{figure}

\textbf{RoboCasa} \cite{nasiriany2024robocasalargescalesimulationeveryday} is a large-scale simulation framework, which provides various tasks in everyday scenarios. Besides the large amounts of tasks, there are extensive intra-task variations in RoboCasa. The variations include scenes, objects, and initial positions of the robot and the objects, while LIBERO does not provide this kind of diversity. As shown in Figure \ref{fig:libero_robocasa_diff}, in the CoffeeServeMug task, the coffee mugs, the coffee machine, and their surroundings are different and indicate different constraints, i.e. the robot can not grasp the mug from left in the left third variation due to the toast machine. This high level of diversity demands strong generalization from the model, which makes this benchmark very challenging.


We evaluate 5 tasks in RoboCasa with 50 human demonstrations for each task. The 5 tasks contain different behaviors, CloseSingleDoor, OpenDrawer, TurnOnStove, CoffeePressButton, and CoffeeServeMug. For training, we train each model for 200 epochs and rollout the models for 50 episodes for each task. We report the success rate over 3 seeds.

\begin{table*}[t!]
\begin{center}
\begin{small}
\begin{sc}
\resizebox{\textwidth}{!}{%
\renewcommand{\arraystretch}{1.2}
\begin{tabular}{ll|cc|cc|cc|cc|cc}
\toprule
&  
& \multicolumn{2}{c|}{\textbf{LIBERO-Spatial}}
& \multicolumn{2}{c|}{\textbf{LIBERO-Object}}
& \multicolumn{2}{c|}{\textbf{LIBERO-Goal}}
& \multicolumn{2}{c|}{\textbf{LIBERO-Long}}
& \multicolumn{2}{c}{\textbf{Avg.}}
\\ 
\midrule
Existing models & Backbones & 20\% & 100\% & 20\% & 100\% & 20\% & 100\% & 20\% & 100\% & 20\% & 100\% \\
\midrule
Diffusion Policy
& Transformer
& -
& $78.3\scriptstyle \pm 1.1$
& -
& $92.5\scriptstyle \pm 0.7$
& -
& $68.3\scriptstyle \pm 1.2$
& -
& $50.5\scriptstyle \pm 1.3$
& -
& $72.4$
\\
Octo
& Transformer
& -
& $78.9\scriptstyle \pm 1.0$
& -
& $85.7\scriptstyle \pm 0.9$
& -
& $84.6\scriptstyle \pm 0.9$
& -
& $51.1\scriptstyle \pm 1.3$
& -
& $75.1$
\\
OpenVLA
& Llama
& -
& $84.7\scriptstyle \pm 0.9$
& -
& $88.4\scriptstyle \pm 0.8$
& -
& $79.2\scriptstyle \pm 1.0$
& -
& $53.7\scriptstyle \pm 1.3$
& -
& $76.5$
\\
MDT
& Transformer
& -
& $78.5 \scriptstyle \pm 1.5$
& -
& $87.5 \scriptstyle \pm 0.9$
& -
& $73.5 \scriptstyle \pm 2.0$
& -
& $64.8 \scriptstyle \pm 0.3$
& -
& $76.1$
\\
MAIL
& Mamba
& $57.6 \scriptstyle \pm 2.2$
& $74.3 \scriptstyle \pm 4.3$
& $78.1 \scriptstyle \pm 4.5$
& $90.1 \scriptstyle \pm 2.0$
& $56.5 \scriptstyle \pm 2.3$
& $81.8 \scriptstyle \pm 1.6$
& \underline{$49.9 \scriptstyle \pm 4.8$}
& $78.6 \scriptstyle \pm 3.0$
& $60.5$
& $83.5$
\\
ATM
& Transformer
& \underline{\boldsymbol{$79.0 \scriptstyle \pm 3.7$}}
& -
& \underline{$81.0 \scriptstyle \pm 2.4$}
& -
& \underline{$58.6 \scriptstyle \pm 4.6$}
& -
& $44.0 \scriptstyle \pm 6.38$
& -
& \underline{$65.6$}
& -
\\
EnerVerse
& Transformer
& -
& \underline{$91.2$}
& -
& \underline{$97.7$}
& -
& \underline{$85.0$}
& -
& \underline{$80.0$}
& -
& \underline{$88.5$}
\\
\midrule
X-BC (ours)\\
\midrule
Dec
& Transformer
& $63.6 \scriptstyle \pm 1.8$
& \underline{$91.7 \scriptstyle \pm 2.7$}
& $89.7 \scriptstyle \pm 1.6$
& $97.4 \scriptstyle \pm 1.0$
& $57.4 \scriptstyle \pm 2.0$
& $87.1 \scriptstyle \pm 4.0$
& $24.3 \scriptstyle \pm 5.9$
& \underline{$79.1 \scriptstyle \pm 1.5$}
& $58.8$
& $88.8$
\\
& Mamba
& $66.2 \scriptstyle \pm 10.7$
& $84.1 \scriptstyle \pm 2.7$
& $72.8 \scriptstyle \pm 25.0$
& \underline{$97.9 \scriptstyle \pm 0.4$}
& $72.5 \scriptstyle \pm 2.4$
& $88.4 \scriptstyle \pm 2.3$
& $34.3 \scriptstyle \pm 1.8$
& $72.1 \scriptstyle \pm 2.9$
& $61.5$
& $85.6$
\\
& xLSTM
& \underline{$72.8 \scriptstyle \pm 3.3$}
& $89.8 \scriptstyle \pm 1.2$
& \underline{$93.5 \scriptstyle \pm 1.2$}
& $96.7 \scriptstyle \pm 0.9$
& \underline{$72.7 \scriptstyle \pm 3.2$}
& \underline{$91.7 \scriptstyle \pm 1.7$}
& \underline{$47.6 \scriptstyle \pm 2.7$}
& $78.6 \scriptstyle \pm 1.1$
& \underline{$71.6$}
& \underline{$89.2$}
\\
\midrule
X-BESO (ours) \\
\midrule
Dec
& Transformer
& $66.5 \scriptstyle \pm 4.1$
& $89.5 \scriptstyle \pm 1.2$
& $90.6 \scriptstyle \pm 1.2$
& $98.6 \scriptstyle \pm 0.2$
& $59.8 \scriptstyle \pm 2.8$
& $85.8 \scriptstyle \pm 0.5$
& $40.5 \scriptstyle \pm 2.2$
& $79.3 \scriptstyle \pm 0.9$
& $64.4$
& $88.3$
\\
& Mamba
& $73.3 \scriptstyle \pm 5.4$
& $92.0 \scriptstyle \pm 1.4$
& \underline{\boldsymbol{$96.6 \scriptstyle \pm 0.3$}}
& \underline{$99.1 \scriptstyle \pm 0.2$}
& $69.4 \scriptstyle \pm 5.6$
& \underline{\boldsymbol{$94.5 \scriptstyle \pm 0.8$}}
& $46.3 \scriptstyle \pm 3.4$
& \underline{\boldsymbol{$85.2 \scriptstyle \pm 1.9$}}
& $71.4$
& \underline{\boldsymbol{$92.7$}}
\\
& xLSTM
& \underline{$74.9 \scriptstyle \pm 3.0$}
& \underline{\boldsymbol{$93.5 \scriptstyle \pm 0.5$}}
& $93.5 \scriptstyle \pm 0.5$
& $98.6 \scriptstyle \pm 0.9$
& \underline{$77.9 \scriptstyle \pm 3.5$}
& $92.9 \scriptstyle \pm 0.6$
& \underline{\boldsymbol{$51.8 \scriptstyle \pm 3.6$}}
& $84.1 \scriptstyle \pm 2.6$
& \underline{$74.5$}
& $92.3$
\\
\midrule
X-RF (ours) \\
\midrule
Dec
& Transformer
& $46.3 \scriptstyle \pm 1.1$
& $87.6 \scriptstyle \pm 2.7$
& $94.3 \scriptstyle \pm 2.0$
& $98.6 \scriptstyle \pm 0.9$
& $32.6 \scriptstyle \pm 1.2$
& $82.4 \scriptstyle \pm 2.2$
& $34.4 \scriptstyle \pm 0.4$
& $80.5 \scriptstyle \pm 1.1$
& $51.9$
& $87.3$
\\
& Mamba
& $63.9 \scriptstyle \pm 4.9$
& $92.6 \scriptstyle \pm 2.0$
& $91.9 \scriptstyle \pm 1.0$
& \underline{\boldsymbol{$99.7 \scriptstyle \pm 0.1$}}
& $55.4 \scriptstyle \pm 2.4$
& $93.7 \scriptstyle \pm 1.8$
& $42.7 \scriptstyle \pm 1.5$
& \underline{$84.5 \scriptstyle \pm 0.6$}
& $63.5$
& \underline{$92.6$}
\\
& xLSTM 
& \underline{$76.9 \scriptstyle \pm 3.8$}
& \underline{$92.9 \scriptstyle \pm 0.1$}
& \underline{$92.9 \scriptstyle \pm 1.1$}
& $98.8 \scriptstyle \pm 0.4$
& \underline{\boldsymbol{$79.2 \scriptstyle \pm 4.5$}}
& \underline{$91.9 \scriptstyle \pm 0.2$}
& \underline{$50.5 \scriptstyle \pm 0.4$}
& $84.3 \scriptstyle \pm 0.6$
& \underline{\boldsymbol{$74.9$}}
& $92.0$
\\
\bottomrule
\end{tabular}
}
\end{sc}
\end{small}
\end{center}
    \caption{Results on LIBERO benchmark with 20\% and 100\% demonstrations, averaged across three seeds. The best overall results are highlighted in bold, with category-specific best results underlined. DEC refers to the Decoder-only architecture.
    }
\label{table:main_results_2d}
\end{table*}

\subsection{Experimental Setup in X-IL}
To ensure a fair comparison, we match the model sizes of Transformer, Mamba, and xLSTM. For both the diffusion policy and flow matching policy, we set the number of sampling steps to 4 in the main experiments. In the LIBERO Benchmark, all models use ResNet-18 for image processing, whereas in the RoboCasa Benchmark, we employ FiLM-ResNet18 for image encoding and an attention-based encoder for point cloud inputs.

\subsection{Baselines}
We additionally report the performance of the following baselines:

\textbf{BC-Transfromer} BC-Transformer is used in RoboCasa \cite{nasiriany2024robocasalargescalesimulationeveryday}. It uses a CLIP model and a ResNet-18 with FilM layers to encode goal instructions and the the image-based observations, respectively.

\textbf{Diffusion Policy} \cite{chi2023diffusion} is a visuomotor policy that optimizes the action distribution iteratively using a conditional denoising diffusion process on a learned gradient field. 

\textbf{Octo} \cite{octo_2023} is an open-source vision-language-action (VLA) model trained on a large-scale dataset. It uses a transformer-based diffusion policy that supports both language and goal image as task input. 

\textbf{OpenVLA} \cite{kim2024openvlaopensourcevisionlanguageactionmodel} is an vision-language-action model based on a much larger model Llama 2 7B. 

\textbf{MDT} \cite{reuss2024multimodaldiffusiontransformerlearning} is a diffusion-based framework that is able to learn versatile behavior from multimodal goal specification including images and languages. 

\textbf{MaIL} \cite{jia2024mailimprovingimitationlearning} uses MAMBA to replace transformer-based backbones in the imitation learning. It demonstrates superior performance compared to transformer-based architectures, especially in the case of small datasets. 

\textbf{ATM} \cite{wen2024anypointtrajectorymodelingpolicy} Any-point Trajectory Modeling (ATM) is a framework learning from video demonstrations. ATM predicts the trajectories of arbitrary points in a video frame using images and language instructions as input. 

\textbf{EnerVerse} \cite{huang2025enerverseenvisioningembodiedfuture} is a framework designed for future space generation regarding robotic manipulation tasks. A policy head is added to the video generator in order to predict the corresponding action simultaneously.

\textbf{3D Diffusion Policy (DP3)} \cite{ze20243d}. DP3
extracts point-wise features from single-view points clouds. Robot actions are generated conditioned on these features and the current robot states.


More detailed description of the policy representations see Appendix \ref{appendix:baselines}.

\begin{table*}[t!]
\begin{center}
\begin{small}
\begin{sc}
\resizebox{\textwidth}{!}{%
\renewcommand{\arraystretch}{1.2}
\begin{tabular}{l|l | l|c|c|c|c|c|c}
\toprule
\textbf{Input Types}
& \textbf{Methods}
& \textbf{Backbones}
& \textbf{CloseSingleDoor}
& \textbf{OpenDrawer}
& \textbf{TurnOnStove}
& \textbf{CoffeePressButton}
& \textbf{CoffeeServeMug}
& \textbf{Average}
\\ 
\midrule 
\multirow{4}{*}{RGB}
& BC-Transformer
& Transformer
& $56.0$
& $42.0$
& $32.0$
& $48.0$
& $22.0$
& $40.0$
\\
\cmidrule{2-9} 
& \multirow{3}{*}{X-BESO (ours)}
& Dec-Transformer
& $72.0 \scriptstyle \pm 1.6$
& $56.7 \scriptstyle \pm 2.5$
& $30.7 \scriptstyle \pm 3.8$
& $58.0 \scriptstyle \pm 1.6$
& $18.0 \scriptstyle \pm 4.3$
& $47.1$
\\
&
& Dec-Mamba
& \underline{$73.3 \scriptstyle \pm 1.9$}
& \underline{\boldsymbol{$68.0 \scriptstyle \pm 2.8$}}
& $28.0 \scriptstyle \pm 3.3$
& \underline{$72.0 \scriptstyle \pm 4.3$}
& $20.7 \scriptstyle \pm 3.4$
& $52.4$
\\
&
& Dec-xLSTM
& $70.0 \scriptstyle \pm 1.6$
& $68.0 \scriptstyle \pm 9.9$
& \underline{$34.7 \scriptstyle \pm 2.5$}
& $64.7 \scriptstyle \pm 0.9$
& \underline{$30.7 \scriptstyle \pm 5.2$}
& \underline{$53.6$}
\\
\midrule
\multirow{5}{*}{Point Cloud}
& 3d Diffusion Policy
& UNet
& $62.0 \scriptstyle \pm 2.8$
& $35.8 \scriptstyle \pm 3.2$
& $29.3 \scriptstyle \pm 0.9$
& \underline{$15.3 \scriptstyle \pm 5.1$}
& \underline{$16.7 \scriptstyle \pm 2.5$}
& $31.8$
\\
\cmidrule{2-9}
& \multirow{3}{*}{X-BESO (ours)}
& Dec-Transformer
& $62.0 \scriptstyle \pm 4.9$
& $24.0 \scriptstyle \pm 1.6$
& $46.7 \scriptstyle \pm 6.2$
& $5.3 \scriptstyle \pm 3.8$
& $5.3 \scriptstyle \pm 3.4$
& $28.7$
\\
&
& Dec-Mamba
& $62.0 \scriptstyle \pm 3.3$
& $34.7 \scriptstyle \pm 11.6$
& \underline{\boldsymbol{$54.0 \scriptstyle \pm 2.8$}}
& $9.3 \scriptstyle \pm 3.4$
& $4.7 \scriptstyle \pm 4.1$
& \underline{$32.9$}
\\
&
& Dec-xLSTM
& \underline{$66.0 \scriptstyle \pm 4.3$}
& \underline{$36.0 \scriptstyle \pm 7.1$}
& $49.3 \scriptstyle \pm 0.9$
& $5.3 \scriptstyle \pm 3.4$
& $7.3 \scriptstyle \pm 5.0$
& $32.8$
\\
\midrule
\multirow{3}{*}{Point Cloud + RGB} 
& \multirow{3}{*}{X-BESO (ours)}
& Dec-Transformer
& $72.0 \scriptstyle \pm 1.6$
& $45.3 \scriptstyle \pm 3.4$
& $37.3 \scriptstyle \pm 5.0$
& $76.0 \scriptstyle \pm 1.6$
& $38.0 \scriptstyle \pm 5.7$
& $53.7$
\\
&
& Dec-Mamba
& $72.7 \scriptstyle \pm 1.9$
& \underline{$53.3 \scriptstyle \pm 0.9$}
& \underline{$44.0 \scriptstyle \pm 1.6$}
& $79.3 \scriptstyle \pm 3.8$
& $44.0 \scriptstyle \pm 4.3$
& $58.7$
\\
&
& Dec-xLSTM
& \underline{\boldsymbol{$74.7 \scriptstyle \pm 2.5$}}
& $50.7 \scriptstyle \pm 6.6$
& $42.7 \scriptstyle \pm 8.2$
& \underline{\boldsymbol{$88.0 \scriptstyle \pm 5.9$}}
& \underline{\boldsymbol{$48.7 \scriptstyle \pm 1.9$}}
& \underline{\boldsymbol{$60.9$}}
\\
\bottomrule
\end{tabular}
}
\end{sc}
\end{small}
\end{center}
    \caption{Results for RoboCasa using different input types with 50 human demonstrations, averaged across three seeds. The best overall results are highlighted in bold, with category-specific best results underlined. DEC refers to the Decoder-only architecture.
    }
\label{table:main_results_robocasa}
\end{table*}

\subsection{Evaluation on Visual Inputs}
\textbf{LIBERO.} We report the main results in Table \ref{table:main_results_2d}. To evaluate our framework on LIBERO, we tested BC, BESO, and RF policies using Decoder-only architectures across Transformer, Mamba, and xLSTM backbones. Our results demonstrate that X-IL achieves state-of-the-art performance, surpassing publicly available models. Specifically, xLSTM shows great potential in both 20\% and 100\% data settings, where it achieves 74.5\% average success and 92.3\% average success respectively. 

\textbf{RoboCasa.} We report the main results in Table \ref{table:main_results_robocasa}. Compared to LIBERO, RoboCasa presents a more challenging benchmark due to its dynamically changing background scenes and object variations across demonstrations and evaluations. We tested X-BESO on five tasks within RoboCasa and observed that our approach outperforms the results reported in the original paper. Specifically, using xLSTM-based models, we achieved a higher average success rate of 53.6\%, compared to 40.0\% of BC-Transformer, demonstrating the effectiveness of our method in handling complex and dynamic environments. Additionally, we observe that Mamba and xLSTM outperform Transformer-based backbones, which is consistent with our findings from LIBERO. This result further highlights the potential of leveraging new sequential models in imitation learning, suggesting that alternative architectures beyond Transformers can offer improved efficiency and performance in complex robotic tasks.


\begin{figure}[h!]
    \centering
    \begin{minipage}{0.23\textwidth}
        \includegraphics[width=\linewidth]{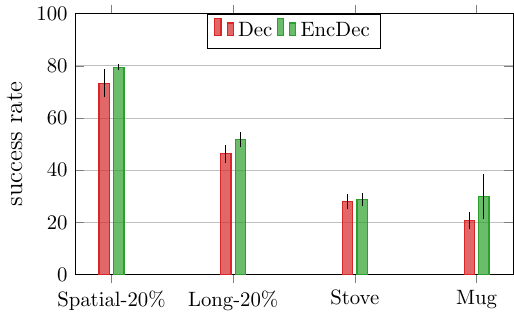}
        \vspace{-0.4cm}
        \subcaption{Mamba}
    \end{minipage}
    \hfill
    \begin{minipage}{0.23\textwidth}
        \includegraphics[width=\linewidth]{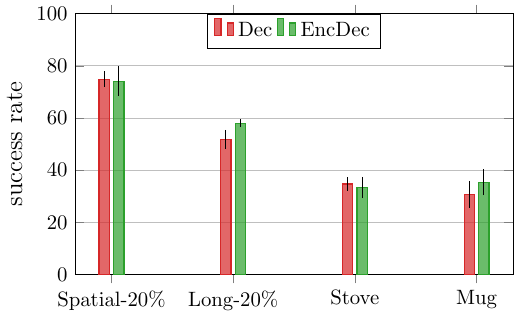}
        \vspace{-0.4cm}
        \subcaption{xLSTM}
    \end{minipage}
    \caption[]{Comparison on different architectures. Dec refers to the Decoder-only model, while EncDec refers to the Encoder-Decoder model.}
    \label{fig:architecture_ablation}
\end{figure}

\subsection{Evaluation on Point Cloud Inputs}
We report the main results in Table \ref{table:main_results_robocasa}.
We evaluate X-BESO using point cloud inputs on RoboCasa and achieve superior results compared to 3D Diffusion Policy. An interesting observation from our results is that point cloud-based inputs do not necessarily outperform RGB-based inputs.

Our analysis suggests that this is due to the complexity of RoboCasa's scenarios, where point clouds are captured from diverse sources, leading to significant information loss during sampling—especially in tasks involving small objects. In such cases, only a sparse set of points remains, limiting the effectiveness of point cloud representations. This highlights the potential benefits of object-centric approaches that focus on preserving critical task-relevant details.

Additionally, we evaluate the performance of combining Point Cloud and RGB inputs. A compact representation is first extracted from the point cloud and then concatenated with the RGB features. Experimental results demonstrate that incorporating both modalities significantly enhances performance, particularly for the xLSTM-based model, which achieves a 60.9\% success rate—compared to 53.6\% with RGB alone and 32.8\% with Point Cloud alone. This highlights the importance of exploring more effective multi-modal fusion strategies to fully leverage the strengths of each modality.


\begin{figure}[h!]
    \centering
    \begin{varwidth}[c]{0.43\textwidth}
        \centering  
        \includegraphics[width=0.43\linewidth]{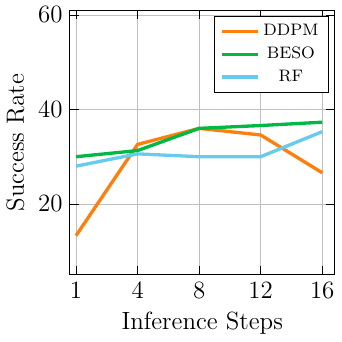}    
    \end{varwidth}\hfill
    \begin{varwidth}[c]{0.53\textwidth}
        \resizebox{0.53\linewidth}{!}{
        \begin{tabular}{p{10mm}|l|l|l}
        \hline
        \textbf{Steps} & DDPM [ms] & BESO [ms] & RF [ms] \\ \hline
        \textbf{1}  & $4.91\scriptstyle \pm 0.05$ & $4.98\scriptstyle \pm 0.09$  & $5.02\scriptstyle \pm 0.03$ \\
        \textbf{4}  & $14.62\scriptstyle \pm 0.28$ &$14.91\scriptstyle \pm 0.29$ & $14.63\scriptstyle \pm 0.34$\\
        \textbf{8}  & $27.39\scriptstyle \pm 0.48$ &$28.00\scriptstyle \pm 0.68$ & $27.65\scriptstyle \pm 0.66$\\
        \textbf{12} & $40.40\scriptstyle \pm 0.80$ &$40.43\scriptstyle \pm 0.75$ & $40.11\scriptstyle \pm 0.74$\\
        \textbf{16} &$53.39\scriptstyle \pm 1.21$ &$53.78\scriptstyle \pm 1.17$ & $52.96\scriptstyle \pm 1.08$\\ \hline
        \end{tabular}
        }
    \end{varwidth}
    \caption{Comparison of different inference steps for DDPM, BESO, and RF. Left: success rate; Right: inference time.}
    \label{fig:diffusion_steps}
\end{figure}

\subsection{Comparison on different architectures}
We conduct experiments on four tasks—Spatial (20\%) and Long (20\%) from LIBERO, as well as TurnOnStove and CoffeeServeMug from RoboCasa—to compare the performance of Decoder-only and Encoder-Decoder architectures. The results, presented in Figure \ref{fig:architecture_ablation}, show that the AdaLN-conditioned Encoder-Decoder architecture achieves superior performance on most tasks, highlighting its effectiveness. Furthermore, by processing observations and actions separately, this design offers more flexibility in choosing different layers for the encoder and decoder, making it more scalable to larger models.

\subsection{Comparison of Diffusion Models Across Varying Inference Steps}
We evaluate Decoder-only xLSTM with DDPM, BESO, and RF on the challenging TurnOnStove task in RoboCasa, comparing performance and inference speed across 1, 4, 8, 12, and 16 inference steps (Figure \ref{fig:diffusion_steps}). DDPM struggles with a single step, while BESO and RF perform well and improve with more steps. Their inference times are similar, and the speed advantage of flow matching is less noticeable due to the lower action dimension.

\subsection{Comparison on different encoders}
\begin{figure}
    \centering
    \begin{minipage}{0.23\textwidth}
        \includegraphics[width=\linewidth]{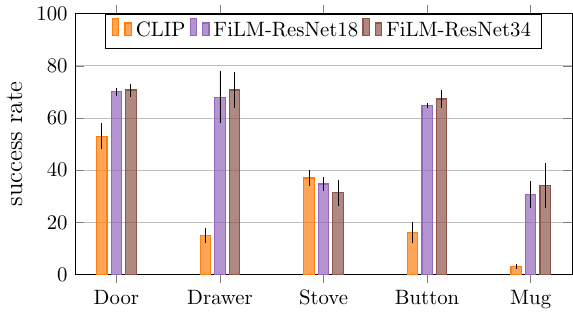}
        \label{fig:img_encoders}
        \vspace{-0.4cm}
        \subcaption{RGB Encoders}
    \end{minipage}
    \hfill
    \begin{minipage}{0.23\textwidth}
        \includegraphics[width=\linewidth]{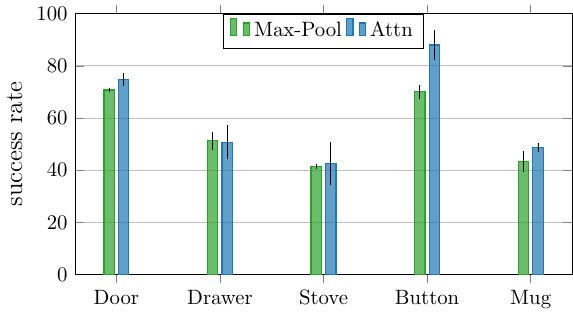}
        \label{fig:img_encoders}
        \vspace{-0.4cm}
        \subcaption{PC Encoders}
    \end{minipage}
    \caption[]{Comparison of different image encoders and Point Cloud (PC) encoders.}
    \label{fig:image_encoders}
\end{figure}

We evaluate different image encoders on the RoboCasa dataset using Dec-xLSTM BESO, comparing FiLM-ResNet18, FiLM-ResNet34, and CLIP (frozen) to assess their impact on performance. We also compare the Max-Pooling and Attention-based Point Cloud encoders. The results are presented in Figure \ref{fig:image_encoders}.

Our findings reveal that despite RoboCasa's requirement for generalization to new scenes and objects, frozen CLIP encoders perform poorly on most tasks. In contrast, fine-tuned ResNet18 and ResNet34 demonstrate strong performance, suggesting that domain adaptation plays a crucial role in achieving effective visual representations for imitation learning. A potential reason for CLIP’s underperformance is 
the significant domain gap between its pretraining data and robot manipulation datasets. This indicates that pretraining on robotics-specific datasets could further improve visual representations for imitation learning.

Compared to the Max-Pooling encoder, the Attention-based encoder shows better performance on most tasks, which indicates that attention can better capture the geometric structures of Point Cloud.






\section{Discussion} \label{sec: discussion}
Here, we list several general observations \textbf{O1)-O4)} that are based on the experiments from Section \ref{sec:experiments}.

\textbf{O1)} Given similar model sizes, Mamba and xLSTM-based policies outperform Transformer-based policies in both RGB inputs and Point Cloud inputs across LIBERO and RoboCasa, demonstrating their potential as viable alternatives to Transformers in imitation learning. 

\textbf{O2)} AdaLN conditioning is a strong method to build Encoder-Decoder architectures in imitation learning and is suitable for all kinds of sequential models. Injecting observation representations to the action decoder through adaLN conditioning could further improve the model's performance compared to the Decoder-only structure.

\textbf{O3)} Point cloud representations do not necessarily outperform RGB-based representations in imitation learning. While previous works have demonstrated strong performance using point cloud inputs, their evaluations were typically conducted on simpler tasks, where the sampled points were highly task-relevant.

For more complex tasks, however, the Furthest Point Sampling (FPS) method distributes points evenly across the scene, often leading to significant information loss—especially for tasks requiring fine-grained object interactions. This suggests that standard point cloud sampling techniques may not always be optimal, and developing task-aware sampling or object-centric approaches could further improve the effectiveness of point cloud representations in imitation learning.

\textbf{O4)} Better strategies for combining point cloud and image representations need further exploration. Our experiments show that simply concatenating point cloud and RGB inputs to the policy improves performance, but the gains are not significant. This suggests that a more structured fusion mechanism is required to fully leverage the complementary nature of these modalities. Finding the right balance between point cloud and image features remains an open challenge.

\textbf{O5)} Our experiments highlight the necessity of robot-specific trained encoders. Comparing fine-tuned FiLM-ResNet with frozen CLIP, we observe that CLIP performs poorly on most tasks. We attribute this to the fact that manipulation tasks require task-specific features, which are not well captured by CLIP’s broad, vision-language pretraining.

\section{Conclusion}
We introduced X-IL, a modular and user-friendly framework for imitation learning, enabling systematic exploration of various policy design choices such as backbones, architectures, and policy representations across multiple modalities. Our framework supports state-of-the-art encoders, efficient sequential models, and multi-modal fusion, providing a unified platform for researchers and practitioners. Through extensive evaluations on LIBERO and RoboCasa, we demonstrate superior performance, improved data efficiency, and better representation learning strategies. Our findings highlight the potential of alternative sequence models, task-adapted encoders, and optimized multi-modal fusion in imitation learning. We hope X-IL serves as a valuable resource for advancing scalable and generalizable policies.

\nocite{langley00}

\bibliography{icml2025}
\bibliographystyle{icml2025}


\newpage
\appendix

\onecolumn

\section{X-IL Details}
\subsection{X-Block}
\label{subsec:x-block}
An X-Block is similar to Diffusion Transformer (DiT) block, but generalized. The key difference is the X-Layer which is capable to plug-in different backbones, including Transformer, Mamba and xLSTM. The Adaptive LayerNorm (AdaLN) is used to make the input tokens conditionally activated on the context. A MLP maps the given context to factors $\alpha$, $\gamma$ and $\beta$. These factors are then applied to scaling and shifting operations in order to manipuate the latent embeddings.

\subsection{Representation Encoders}
\label{subsec:repr-encoders}
\subsubsection{RGB}

\begin{itemize}
    \item \textbf{ResNet} ResNet-18 with a latent dimension of 512 is used in this paper. 
    \item \textbf{FiLM-ResNet} FilM introduces a general conditioning layer for visual reasoning tasks. ResNet-18 with a latent dimension of 512 is used for FilM-ResNet. The FilM has a condition dimension of 512.
    
\end{itemize}

\subsubsection{Point Cloud}
\begin{figure}[h]
        \centering

        \begin{subfigure}{0.35\textwidth}        
        \resizebox{\textwidth}{!}{\includegraphics{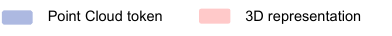}}%
        \end{subfigure}
        
        \begin{subfigure}{0.2\textwidth}        
        \resizebox{\textwidth}{!}{\includegraphics{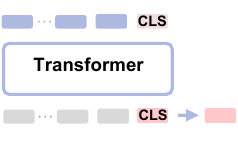}}%
        \caption{Attention}
        \label{fig:pc_encoders_attention}
        \end{subfigure}
        \hspace{1cm}
        \begin{subfigure}{0.2\textwidth}        
        \resizebox{\textwidth}{!}{\includegraphics{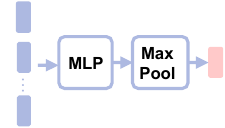}}%
        \caption{MLP with max pooling}
        \label{fig:pc_encoders_mlp}
        \end{subfigure}

        \caption{Point cloud encoding.}
\end{figure}

    \begin{itemize}
    \item \textbf{Attention} We use a 4-layer self-attention Transformer to encode the point cloud tokens. A CLS token is used to capture the whole geometric representation. See Figure \ref{fig:pc_encoders_attention}. 
    \item \textbf{MLP with MaxPooling} We use the same point cloud encoder in 3D Diffusion Policy \cite{Ze2024DP3}, with a 3-layer MLP and Max Pooling to get the compact 3d representations. See Figure \ref{fig:pc_encoders_mlp}. 
    \end{itemize}

\subsubsection{Language}
We used pre-trained CLIP to encode language. CLIP is a vision-language model that aligns the latent embeddings of visual and language inputs. Pre-trained model ViT-B/32 is used as language encoder. We freeze the weights of both models during the training.

\subsection{Backbone Details}
Hyperparameters of three different backbones, i.e., Transformer, Mamba, and xLSTM, used in different policies are shown in the Table \ref{table:appendix_backbone_transformer}, \ref{table:appendix_backbone_mamba} and \ref{table:appendix_backbone_xlstm}. These hyperparameters are not tuned for a specific task but are used for all tasks reported in this paper.

\newcommand{\encoder}[1]{\textcolor{blue}{#1}}
\newcommand{\decoder}[1]{\textcolor{violet}{#1}}

\begin{table}[h]
\begin{center}
\begin{small}
\begin{sc}
\resizebox{0.6\linewidth}{!}{%
\begin{tabular}{l|c|c|c|c|c|c}
\toprule
\textbf{Policy}
& \multicolumn{2}{c|}{DDPM}
& \multicolumn{2}{c|}{BESO}
& \multicolumn{2}{c}{FM}
\vspace{3pt}
\\ 
\textbf{Transformer}
& Dec
& \encoder{Enc}-\decoder{Dec}
& Dec
& \encoder{Enc}-\decoder{Dec}
& Dec
& \encoder{Enc}-\decoder{Dec}
\\ 
\midrule
\midrule
\textbf{Layers}
& 6
& \encoder{4} - \decoder{6}
& 6
& \encoder{4} - \decoder{6}
& 6
& \encoder{4} - \decoder{6}
\\
\midrule
\textbf{Heads}
& 8
& 8
& 8
& 8
& 8
& 8
\\
\midrule
\textbf{Embeddings}
& 512
& 512
& 512
& 512
& 512
& 512
\\
\midrule
\textbf{Batch Size}
& 256
& 256
& 256
& 256
& 256
& 256
\\
\midrule
\textbf{Learning rate}
& 1e-4
& 1e-4
& 1e-4
& 1e-4
& 1e-4
& 1e-4
\\
\midrule
\textbf{Optimizer}
& Adam
& Adam
& AdamW
& AdamW
& AdamW
& AdamW
\\
\bottomrule
\end{tabular}
}
\end{sc}
\end{small}
\end{center}
    \caption{Hyperparameters of Transformer}
\label{table:appendix_backbone_transformer}
\end{table}
\begin{table}[h]
\begin{center}
\begin{small}
\begin{sc}
\resizebox{0.6\linewidth}{!}{%
\begin{tabular}{l|c|c|c|c|c|c}
\toprule
\textbf{Policy}
& \multicolumn{2}{c|}{DDPM}
& \multicolumn{2}{c|}{BESO}
& \multicolumn{2}{c}{FM}
\vspace{3pt}
\\ 
\textbf{Mamba}
& Dec
& \encoder{Enc}-\decoder{Dec}
& Dec
& \encoder{Enc}-\decoder{Dec}
& Dec
& \encoder{Enc}-\decoder{Dec}
\\ 
\midrule
\midrule
\textbf{Layers}
& 8
& \encoder{4} - \decoder{8}
& 8
& \encoder{8} - \decoder{8}
& 8
& \encoder{4} - \decoder{8}
\\
\midrule
\textbf{Embeddings}
& 512
& 512
& 512
& 512
& 512
& 512
\\
\midrule
\textbf{Batch Size}
& 256
& 256
& 256
& 256
& 256
& 256
\\
\midrule
\textbf{Learning rate}
& 1e-4
& 1e-4
& 1e-4
& 1e-4
& 1e-4
& 1e-4
\\
\midrule
\textbf{Optimizer}
& Adam
& Adam
& AdamW
& AdamW
& AdamW
& AdamW
\\
\bottomrule
\end{tabular}
}
\end{sc}
\end{small}
\end{center}
    \caption{Hyperparameters of Mamba}
\label{table:appendix_backbone_mamba}
\end{table}
\begin{table}[h]
\begin{center}
\begin{small}
\begin{sc}
\resizebox{0.6\linewidth}{!}{%
\begin{tabular}{l|c|c|c|c|c|c}
\toprule
\textbf{Policy}
& \multicolumn{2}{c|}{DDPM}
& \multicolumn{2}{c|}{BESO}
& \multicolumn{2}{c}{FM}
\vspace{3pt}
\\ 
\textbf{xLSTM}
& Dec
& \encoder{Enc}-\decoder{Dec}
& Dec
& \encoder{Enc}-\decoder{Dec}
& Dec
& \encoder{Enc}-\decoder{Dec}
\\ 
\midrule
\midrule
\textbf{Blocks}
& 8
& \encoder{4} - \decoder{8}
& 8
& \encoder{8} - \decoder{8}
& 8
& \encoder{4} - \decoder{8}
\\
\midrule
\textbf{Embeddings}
& 512
& 512
& 512
& 512
& 512
& 512
\\
\midrule
\textbf{Batch Size}
& 256
& 256
& 256
& 256
& 256
& 256
\\
\midrule
\textbf{Learning rate}
& 1e-4
& 1e-4
& 1e-4
& 1e-4
& 1e-4
& 1e-4
\\
\midrule
\textbf{Optimizer}
& Adam
& Adam
& AdamW
& AdamW
& AdamW
& AdamW
\\
\bottomrule
\end{tabular}
}
\end{sc}
\end{small}
\end{center}
    \caption{Hyperparameters of xLSTM}
\label{table:appendix_backbone_xlstm}
\end{table}

\newpage


\subsection{Policies}
\label{appendix:policy}
\textbf{Behavior Cloning} Behavior cloning (BC) assumes a Gaussian distribution as policy representation and has therefore limited model capacity. Maximizing the likelihood of the policy results in a mean squared error (MSE) minimization between ground truth and predicted actions. Due to its simplicity, BC is often used as a default naive baseline for comparing imitation learning policies.  

\textbf{Diffusion-Based Policies}  
Denoising diffusion probabilistic models (DDPM) \cite{ho2020denoising} are a popular choice for policy representation due to their simplicity and minimal design choices compared to advanced models like BESO \cite{reuss2023goal}. BESO, based on a continuous-time diffusion framework, allows for varying diffusion steps during training and inference, as well as diverse sampling techniques such as DDIM. Despite these differences, both DDPM and BESO rely on regression losses, either learning a score function or a denoising model. Our framework supports both DDPM-style policies and continuous-time BESO-style policies.  

\textbf{Flow-Based Policies}  
Continuous-time normalizing flows trained via flow matching \cite{lipman2022flow} have recently gained a lot of attention and are also suitable as policy representation. These methods, often referred to as rectified flows (RF) \cite{liu2022flow} or stochastic interpolants \cite{albergo2022building}, are fully supported in our framework.

\section{Additional Task Details}

\subsection{LIBERO Benchmark}

The LIBERO Benchmark comprises five distinct task suites: \textit{LIBERO-Spatial}, \textit{LIBERO-Object}, \textit{LIBERO-Goal}, \textit{LIBERO-Long}, and \textit{LIBERO-90}. Each suite is designed to evaluate distinct dimensions of robotic learning and manipulation capabilities:

\begin{itemize}
    \item \textit{LIBERO-Spatial} evaluates spatial reasoning precision through tasks requiring differentiation of identical objects (e.g., bowls) based solely on their relational positioning (e.g., placement relative to plates or other objects).
    \item \textit{LIBERO-Object} tests object-centric manipulation by requiring precise recognition and interaction with unique objects (e.g., pick-and-place tasks) in each trial, emphasizing perceptual discrimination.
    \item \textit{LIBERO-Goal} assesses goal-conditioned adaptability by defining distinct objectives (e.g., placing objects in varying sequences or configurations) within fixed object-layout environments, necessitating dynamic behavioral adjustments.
    \item \textit{LIBERO-Long} challenges long-term planning and endurance with multi-stage tasks demanding sustained execution, error mitigation, and coordination across extended timelines.
    \item \textit{LIBERO-90} serves as a generalization benchmark, aggregating 90 short-horizon tasks across heterogeneous settings to evaluate robustness to variability in objects, layouts, and objectives.
\end{itemize}

\begin{figure*}[h!]
    \centering
    \includegraphics[width=0.85\linewidth]{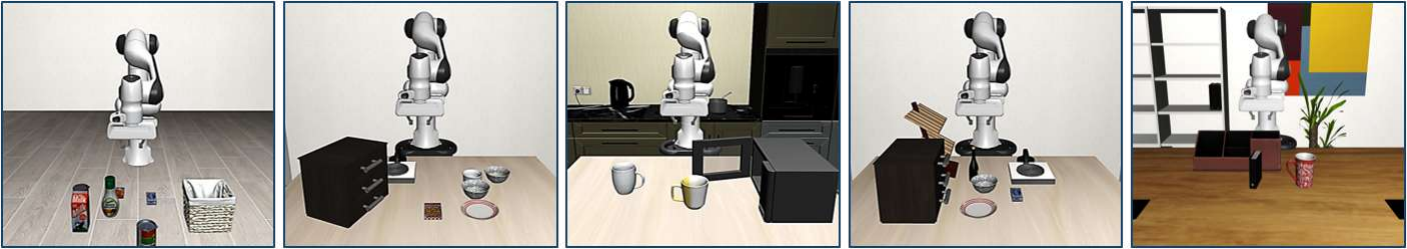}
    \caption{LIBERO benchmark suites with total 130 tasks in five different scenes}
    \label{fig:libero}
\end{figure*}

With the exception of \textit{LIBERO-90}, which includes a diverse collection of 90 tasks, the remaining suites each consist of 10 tasks accompanied by 50 demonstrations per task. For the purposes of training and evaluation efficiency, our study focuses on the first four suites (LIBERO-Spatial, LIBERO-Object, LIBERO-Goal, and LIBERO-Long).

\subsection{Robocasa Benchmark}

RoboCasa is a large-scale simulation framework developed to train generalist robots in realistic and diverse home environments, with a particular focus on kitchen scenarios. The benchmark comprises 100 tasks, including 25 atomic tasks with 50 human demonstrations and 75 composite tasks with auto-generated demonstrations. These tasks are centered around eight fundamental robotic skills relevant to real-world home environments: (1) pick-and-place, (2) opening and closing doors, (3) opening and closing drawers, (4) twisting knobs, (5) turning levers, (6) pressing buttons, (7) insertion, and (8) navigation.


To comprehensively evaluate our method, we selected five tasks from the atomic tasks, each representing a distinct skill:

\begin{itemize}
    \item \textit{Close Single Door} :  Opening and closing doors
    \item \textit{Open Drawer} :  Opening and closing drawers
    \item \textit{Turn on Stove} :  Twisting knobs
    \item \textit{Coffee Press Button} :  Pressing buttons
    \item \textit{Coffee Serve Mug} :  Insertion
\end{itemize}

One of the key advantages of RoboCasa is its provision of both image-based (RGB) and 3D observations, enabling evaluations across different sensing modalities. Accordingly, we assessed our method using both RGB and 3D observations.





\section{Additional Baselines Details}
\subsection{Baselines}
\label{appendix:baselines}

\textbf{Diffusion Policy} \cite{chi2023diffusion} is a visuomotor policy that optimizes the action distribution iteratively using a conditional denoising diffusion process on a learned gradient field. It demonstrates the capability to capture the multi-modal action distributions. Incorporating techniques such as receding horizon control and visual conditioning, the learned visuomotor policy can be deployed to real-world embodiments.

\textbf{Octo} \cite{octo_2023} is an open-source vision-language-action (VLA) model. It uses a transformer-based diffusion policy that supports both language and goal image as task input. The policy is trained on a large-scale dataset and can be deployed to various embodiments.

\textbf{OpenVLA} \cite{kim2024openvlaopensourcevisionlanguageactionmodel}is another open-source vision-language-action model. Different from Octo, OpenVLA is based on a much larger model Llama 2 7B. It uses 256 reserved tokens for action values, providing higher resolution for robot control signals.

\textbf{MDT} \cite{reuss2024multimodaldiffusiontransformerlearning} is a diffusion-based framework that is able to learn versatile behavior from multimodal goal specification including images and languages. It introduces latent goal representations by aligning the latent embeddings of the goal image and the language instructions of the same tasks. This alignment is especially beneficial when there are few language annotations in the dataset. 

\textbf{MaIL} \cite{jia2024mailimprovingimitationlearning} uses MAMBA to replace transformer-based backbones in the imitation learning. It demonstrates superior performance compared to transformer-based architectures, especially in the case of small datasets. 

\textbf{ATM} \cite{wen2024anypointtrajectorymodelingpolicy} Any-point Trajectory Modeling (ATM) is a framework learning from video demonstrations. ATM predicts the trajectories of arbitrary points in a video frame using images and language instructions as input. The robot action is derived from trajectory predictions and the current video frame using MLP. 

\textbf{EnerVerse} \cite{huang2025enerverseenvisioningembodiedfuture} is a framework designed for future space generation regarding robotic manipulation tasks. This framework uses Free Anchor View (FAV) and 4D Gaussian Splatting (4DGS) to generate the next frames of the videos in manipulation scenarios. A policy head is added to the video generator in order to predict the corresponding action simultaneously. EnverVerse shows improved performance in long-horizon manipulation tasks.


\end{document}